\documentclass{article}

\PassOptionsToPackage{numbers, compress}{natbib}
\usepackage[final]{neurips_2020}

\usepackage[utf8]{inputenc} % allow utf-8 input
\usepackage[T1]{fontenc}    % use 8-bit T1 fonts
\usepackage{hyperref}       % hyperlinks
\usepackage{url}            % simple URL typesetting
\usepackage{booktabs}       % professional-quality tables
\usepackage{amsfonts}       % blackboard math symbols
\usepackage{nicefrac}       % compact symbols for 1/2, etc.
\usepackage{microtype}      % microtypography

\usepackage{multicol}
\usepackage{multirow}
\usepackage{tipa}
\usepackage[force]{covington}

\title{Predicting metrical patterns in Spanish poetry with language models}

\author{Javier de la Rosa \\
  UNED \\
  Madrid, Spain \\
  {\tt versae@linhd.uned.es} \\
\And
  Salvador Ros \\
  UNED \\
  Madrid, Spain \\
  {\tt sros@scc.uned.es} \\
\And
 Elena González-Blanco \\
 IE University \\
 Madrid, Spain \\
 {\tt egonzalezblanco@faculty.ie.edu} \\
}

\begin{document}

\maketitle

\begin{abstract}
  In this paper, we compare automated metrical pattern identification systems available for Spanish against extensive experiments done by fine-tuning language models trained on the same task. Despite being initially conceived as a model suitable for semantic tasks, our results suggest that BERT-based models retain enough structural information to perform reasonably well for Spanish scansion.
\end{abstract}

\section{Introduction}
\label{intro}
We can consider the metre of a verse as a sequence of stressed (strong) and unstressed (weak) syllables, which are sometimes denoted with the plus symbol `$+$' for stressed syllables and the minus `$-$' for the unstressed ones. Example \ref{example.1} shows an hendecasyllabic Spanish verse and the resulting metrical pattern after applying rhetorical figures that might shrink (synalepha) or expand (syneresis) its length. The stress of the last word also affects the metrical length in Spanish poetry. The identification of metrical patterns is part of a larger procedure for the scansion of a poem.

\begin{example}\label{example.1}
\textit{cubra de nieve la hermosa cumbre}\footnote{{"[It]} covers with snow the beautiful summit."}\\
\textit{\textbf{cu}-bra-de-\textbf{nie}-ve-la-her-\textbf{mo}-sa-\textbf{cum}-bre} \\
$+--+---+-+-$ 11 \\
(Garcilaso de la Vega) \\
\end{example}

Research has shown that neural models implicitly encode linguistic features ranging from token labeling to different kinds of segmentation \cite{liu2019linguistic}. There is also evidence that language models and embeddings are able to capture not only semantic and syntactic properties but structural information, as shown by \citet{hewitt2019structural} in their work with structural probes for extracting syntax trees, and \citet{conneau2018you} approximating the length in words of a sentence by its vector.

On the other hand, the earliest computational approach to Spanish scansion was introduced by \citet{gervas2000logic}, whose tool uses DCG (Definite Clause Grammars) to model word syllabification as well as additional predicates to define synalepha, syllable count and rhyme. More recently, the ADSO Scansion system introduced by \citet{navarro2017metrical} first applies part of speech (PoS) tags to the words of every line in a poem to analyze hendecasyllables. Rantanplan \cite{delarosa2020rantanplan} employs a similar system but focused on accuracy and speed and achieves state of the art results. The only neural approach was explored by \citet{agirrezabal2016zeuscansion,agirrezabal2017comparison}, who used bi-LSTM neural networks and CRF's to automatically scan poetry in three languages (i.e. English, Spanish and Basque).

% \section{Materials and Methods}
% This paper evaluates BERT's multilingual model and compares it to other current methods for the metrical scansion of poetry in 3 languages. This section describes the data involved in the experiments and the systems evaluated. We also introduce the experimental setup used.

\section{Experimental Design and Results}
As a corpus for Spanish, we decided to use the \textit{Corpus de Sonetos de Siglo de Oro} ``Golden Age Spanish corpus'' \cite{navarro2016metrical}. Although most of the poems included were annotated automatically, it includes 730 poems with manually annotated metrical information, consisting of over 71,000 lines. From this corpus, a subset of 100 poems was used to evaluate \citet{navarro2017metrical}'s automated scansion system. We used only this subset, which we consider a gold standard for Spanish metrical annotation applied to hendecasyllables. This corpus, annotated in TEI-XML, contains sonnets from canonical Golden Age Spanish authors (16th and 17th centuries), featuring only hendecasyllabic verses.

Our downstream task is defined as metrical pattern prediction, that is, given a raw string of text representing a line of verse of a poem, a model is expected to predict a string of $+$ and $-$ symbols representing the stress of each syllable after any rhetorical device has been applied. We split the corpus into training (6,558 lines), evaluation (2,187 lines), and test (1,401 lines) sets. Using the training and evaluation sets, we fine-tuned a series of BERT-based models and a baseline classifier using fasttext \cite{bojanowski2017enriching} with multi-class binary classification. Specifically, we added a fully connected layer to predict the presence or absence of stress in each of the 11 positions of the hendecasyllabic verses in the corpus. We used the language Python 3, the library PyTorch, and the framework Transformers \cite{wolf2019huggingface}. We pre-processed all texts removing duplicate verses and punctuation marks since they are irrelevant for metrical purposes. We trained the models using an AdamW optimiser with a learning rate set to 3e-5 \cite{devlin2018bert}, gradient clipping of 1.0, early stopping set at 5, and for 100 epochs. Training was run on a 30GB of RAM memory with one NVIDIA Tesla V100 GPU with 16GB of memory running on Debian 10. The maximum sequence length was set at 32 tokens and the batch size at 16.

Instead of calculating standard measures for multi-class classification, which in our case would produce per-syllable metrics, we decided to consider as a correct prediction only when all the individual syllable predictors were correct per-line. Therefore, we are reporting accuracy expressed as a percentage of the testing set.

\begin{table} [htbp]
\begin{center}
\begin{tabular} {lcc}
  \hline\rule{-2pt}{15pt}
  & \multicolumn{2}{c}{\bf Accuracy} \\
  {\bf Method} & {\bf 10 Epochs} & {\bf 100 Epochs}\\
  \hline
    Baseline (fasttext) & 10.89 & 11.20 \\
  \hline
    Multilingual BERT & 73.45 & 85.15 \\
    RoBERTa (base) & 76.52 & 87.37 \\
    \textbf{RoBERTa (large)} & \textbf{87.58} & \textbf{93.08} \\
    XLM RoBERTa (base) & 61.53 & 82.16 \\
    XLM RoBERTa (large) & 85.15 & 91.51 \\
    Spanish BERT & 47.47 & 72.73 \\
  \hline\rule{-4pt}{10pt}
    \textbf{Rantanplan \cite{delarosa2020rantanplan}} (SOTA) & \multicolumn{2}{c}{\bf 96.23} \\

\hline
\end{tabular}
\end{center}
\caption{\label{table.2}Accuracy of the different methods. Best scores in bold}
\end{table}

As illustrated in Table \ref{table.2}, results show that BERT-based models are capable of predicting metrical patterns 9 times better tha baseline and reasonably well soon after training starts, performing close to state of the art rule-based systems for longer training times.

\section{Conclusions and Further Work}
In this paper we have evaluated the capabilities of BERT-based models when trained on the task of predicting the metrical pattern of a verse. Under the assumption that transformed-based models were capable of performing tasks of structural nature beyond those of the semantic kind, we show that BERT models perform reasonably well for Spanish. These results suggest that extended pre-training on poetic corpora and hyperparameters search could further improve on the task of metrical pattern prediction. Moreover, there seems to be advantages on the use of multilingual models over the monolingual Spanish BERT, which suggests that important information on the prosody and syllabification of languages is somehow shared for the task of stress assignment.

\begin{ack}
This research was supported by the project Poetry Standardization and Linked Open Data (POSTDATA) (ERC-2015-STG-679528) obtained by Elena González-Blanco and funded by an European Research Council (\url{https://erc.europa.eu}) Starting Grant under the Horizon2020 Program of the European Union.
\end{ack}

\medskip

% \small
\bibliography{lxai_neurips_2020}

\end{document}